\documentclass[conference]{IEEEtran}
\IEEEoverridecommandlockouts
\usepackage{cite}
\usepackage{amsmath,amssymb,amsfonts}
\usepackage{graphicx}
\usepackage{textcomp}
\usepackage{flushend}
\usepackage{xcolor}
\usepackage{balance}
\usepackage{multirow}
\usepackage{subcaption}
\usepackage{tabularx}
\usepackage{array}
\usepackage{algorithm}
\usepackage{algpseudocode}
\usepackage{hyperref}
\usepackage{orcidlink}
\def\BibTeX{{\rm B\kern-.05em{\sc i\kern-.025em b}\kern-.08em
    T\kern-.1667em\lower.7ex\hbox{E}\kern-.125emX}}

\begin{document}

\title{AI-Driven Digital Twin Framework for Warehouse Logistics Automation Optimization}

\author{\IEEEauthorblockN{$^{\ddagger}$Md Mahinur Alam*\textsuperscript{\orcidlink{0009-0006-5453-443X}}, $^{\dagger}$Kanita Jerin Tanha*\textsuperscript{\orcidlink{0009-0006-5453-443X}}, $^{\mathsection}$Md Tanjin Alam\textsuperscript{\orcidlink{}}, $^{\varphi}$Shadman Sakib\textsuperscript{\orcidlink{0009-0006-5453-443X}}, $^{\ddagger}$Min Gyu Kim\textsuperscript{\orcidlink{}}} 
\IEEEauthorblockA{
$^{\dagger}$Kumoh~National~Institute~of~Technology,~Gumi,~South~Korea, \\
$^{\ddagger}$Georgia Southern University, GA, USA, \\
$^{\varphi}$University of Maryland, Baltimore County, MD, USA, $^{\mathsection}$VosuAI,\\
kanitajerin17@kumoh.ac.kr, piash.tanjin@gmail.com,
(malam4, ssakib1)@umbc.edu,
mkim@georgiasouthern.edu}
\thanks{*Equal Contribution}
}

\maketitle

\begin{abstract}
In recent years, warehouse logistics optimization has increasingly focused on intelligent, data-driven systems to enhance operational efficiency. Conventional static product placement strategies are being progressively replaced by dynamic slotting approaches that can adapt to real-time variations in warehouse conditions. In this work, we present a novel framework that employs a reinforcement learning agent integrated within a FlexSim-based Digital Twin environment for real-time product placement optimization. The proposed agent continuously learns from streaming operational data, enabling it to reduce picking time while improving overall system throughput. The simulation environment models a four-zone warehouse layout with forklift-based material handling and leverages the Asset Administration Shell framework to ensure seamless interaction between the learning agent, physical asset, and the digital twin. Experimental evaluation demonstrates substantial performance gains, with the proposed approach achieving a 29.66\% reduction in average picking time, 35.42\% increase in throughput, and 18.54\% decrease in travel distance.

\end{abstract}
\begin{IEEEkeywords}
Automation, digital twin, reinforcement learning (RL), warehouse logistics.	 
\end{IEEEkeywords}

\section{Introduction}\label{sec1}
Warehouse logistics increasingly face volatile demand, labor shortages, and tight service-level constraints, making static slotting and rule-based placement strategies inadequate for sustaining low picking time and high throughput. Digital twin (DT) technology addresses this limitation by providing a high-fidelity, continuously updatable virtual replica of warehouse operations that supports real-time monitoring and safe evaluation of alternative decisions without interrupting physical workflows. To ensure interoperable and consistent interactions across heterogeneous assets (e.g., racks, transporters, conveyors), the Asset Administration Shell (AAS) \cite{tanha2025flexsim} further standardizes asset identity, state, and control interfaces, enabling reliable data exchange between the DT environment and intelligent optimization modules \cite{alam2026ai}.

\begin{figure} [h]
\centering
\includegraphics[width=9cm, height = 5cm]{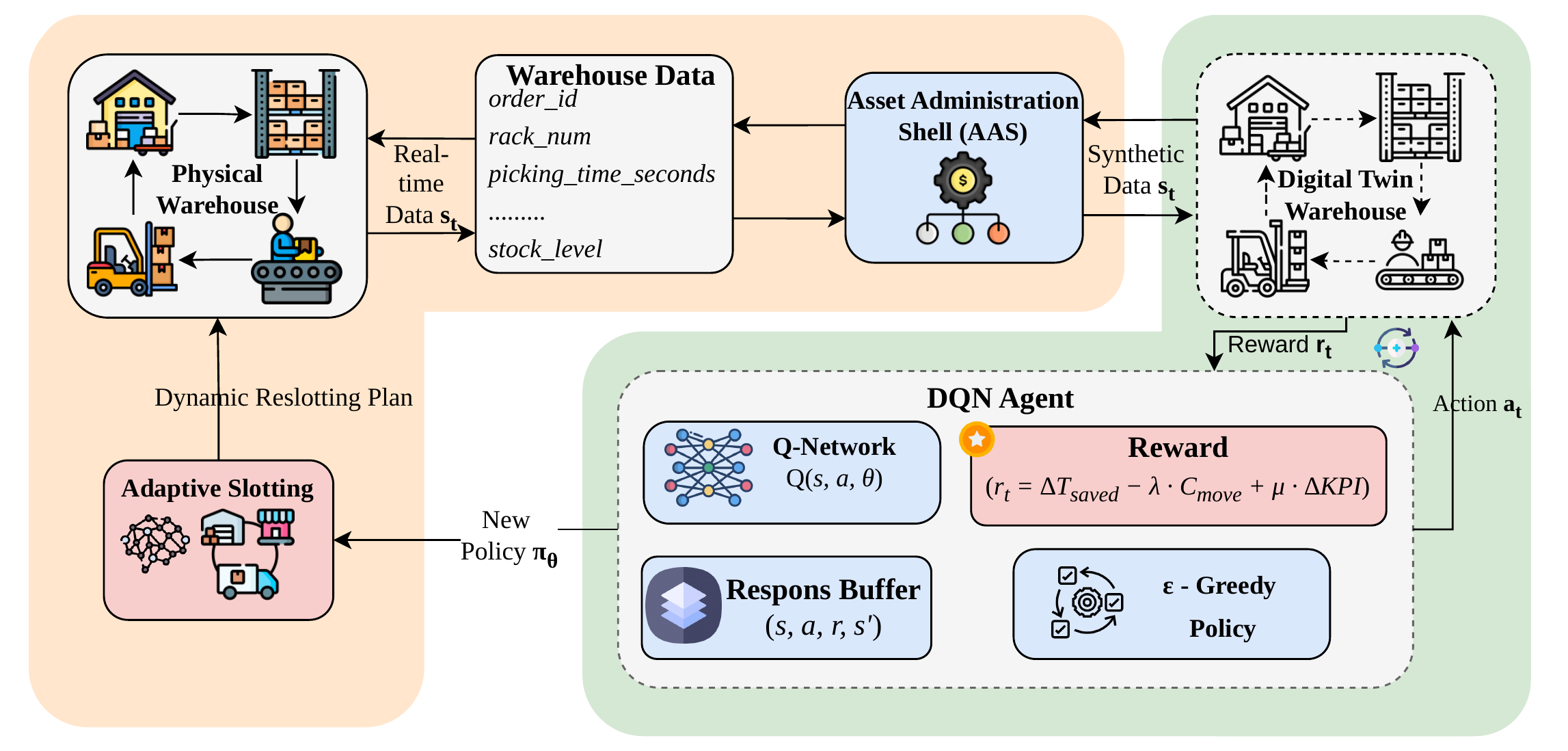}
\caption{Proposed AI-Driven Digital Twin Framework for Smart Warehouse Logistics, Physical-to-Digital Synchronization via Asset Administration Shell with Reinforcement Learning Based Dynamic Reslotting Optimization.}
\label{fig1}
\end{figure}

Dynamic warehouse optimization has been explored through
operations research, rule-based heuristics, and learning-based
methods \cite{jia2017dynamic}. Classical approaches include
MILP-based hierarchical frameworks for multi-AGV scheduling,
which achieve notable reductions in task completion time, but
depend on static formulations that limit real-time adaptability
\cite{hu2023dynamic}. Similarly, genetic-algorithm
variants such as ARGA \cite{cheng2023efficient} and multi-agent
pathfinding decomposition methods \cite{li2023double} offer
strong coordination capabilities, but incur significant
computational overhead and do not adapt dynamically to
evolving warehouse conditions.

Reinforcement learning (RL) has gained prominence as a more
adaptive alternative for storage assignment and routing
\cite{islam2021novel}. Proximal Policy Optimization (PPO) has
demonstrated a 6.3\% reduction in transport costs through
adaptive slot selection, outperforming classical ABC heuristics
\cite{waubert2022dynamic}. Deep Q-Network (DQN) methods have
been applied to task assignment and shelf reallocation, reducing
travel time and improving throughput over heuristic-based
planning \cite{wu2024deep}. Other deep RL approaches using
The Asprilo environment has minimized travel distance,
outperforming GA, ACO, and VRP heuristics by 1.76\% across
various warehouse layouts \cite{li2025reinforcement}. Despite
this advances, most RL-based solutions operate in isolated
simulation environments without integration into high-fidelity
digital twin systems, limiting their ability to continuously
reflect real warehouse states and hinder practical deployment.

To bridge this gap, we present an integrated framework that
couples a DQN-based RL agent with a FlexSim-based Digital Twin
through an AAS middleware layer, enabling real-time
perception--action loops for dynamic slotting optimization. The
key contributions of this paper are: \textbf{(1)} We propose a DQN-based dynamic slotting framework that learns demand-aware zone assignment policies from real-time warehouse states, integrating action masking, experience replay, and 
$\epsilon$-greedy exploration to jointly optimize picking time, travel distance, and throughput. \textbf{(2)} We design a hybrid CNN–BiLSTM–Attention Q-network architecture that captures spatial feature correlations and temporal dependencies in warehouse state sequences, and validate its effectiveness through ablation studies against alternative architectures.
\textbf{(3)} We implement a high-fidelity FlexSim-based Digital Twin integrated with the AAS, enabling standardized asset representation, seamless bidirectional data exchange, and safe shadow evaluation of slotting policies prior to physical deployment.

\begin{figure}[h]
	\centering
    \includegraphics[width=1.00\linewidth]{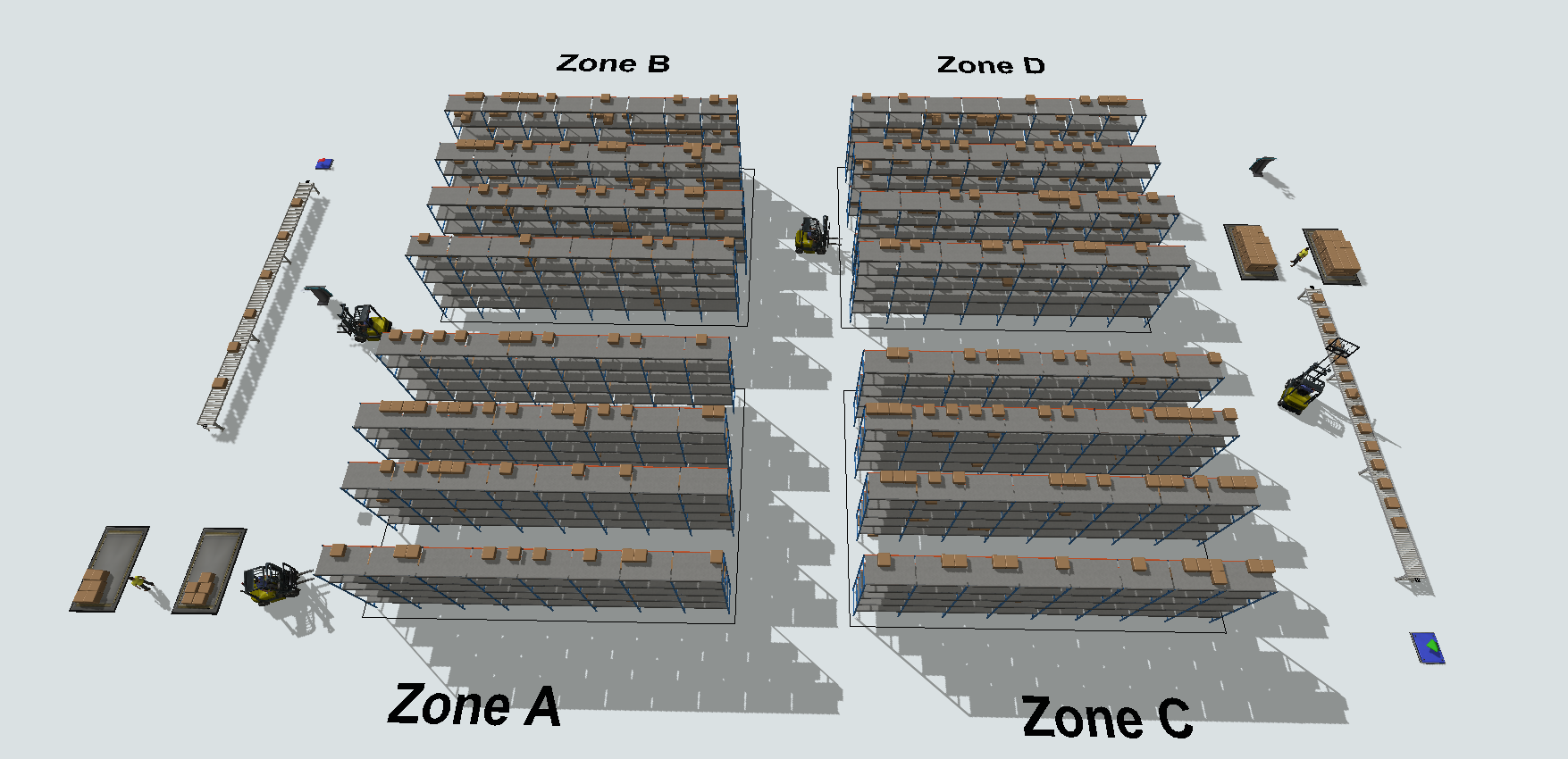}
	\caption {Digital Twin Warehouse Logistics Simulation in FlexSim Four Zone Layout with Forklift-Based Product Handling.}
\label{demo}
\end{figure}

\section{Proposed Framework}
\label{sec3}

The proposed framework tightly couples a reinforcement learning agent with a FlexSim-based Digital Twin through an Asset Administration Shell middleware layer, as illustrated in Fig.~\ref{fig1}. IoT-instrumented physical warehouse assets are mirrored digitally via AAS, which standardizes asset identification, state representation, and control interfaces, ensuring bidirectional synchronization between the physical facility and its simulated counterpart. The RL agent operates exclusively within the DT environment, issuing dynamic slotting decisions that are evaluated in simulation before deployment as adaptive reslotting plans.

\subsection{RL Agent Formulation}

At each decision step $t$, the warehouse state is encoded as
$s_t = [\text{order\_id},\ \text{rack\_num},\ \text{pick\_time},\
\ldots,\ \text{stock\_level}]$, and the agent selects a zone
assignment $a_t \in \{A, B, C, D\}$ for a candidate item. Invalid actions are masked at each step by setting their Q-values to $-\infty$ before action selection, preventing
assignments to zones that have reached rack capacity or
whose current stock levels exceed a predefined threshold,
thereby constraining exploration to operationally feasible
placements. The scalar reward $r_t = \Delta T_{\text{saved}} - \lambda \cdot
C_{\text{move}} + \mu \cdot \Delta\text{KPI}$ balances picking
time improvement $\Delta T_{\text{saved}}$, relocation cost
$C_{\text{move}}$, and temporal change in zone-level performance
$\Delta\text{KPI} = \text{KPI}_{\text{zone}}(t{+}1) -
\text{KPI}_{\text{zone}}(t)$, where
$\text{KPI}_{\text{item}} = \sum_{i} w_i \tilde{x}_i$ aggregates
min-max normalized fulfillment rate ($w_1{=}0.35$), stockout rate
($w_2{=}0.25$), turnover ($w_3{=}0.25$), and holding cost
($w_4{=}0.15$). We set $\lambda{=}0.3$ and $\mu{=}0.2$ via grid
search over $\{0.1, 0.2, 0.3, 0.5\}$, prioritizing picking time
as the dominant signal. The agent remains robust across moderate
ranges; performance degrades only when relocation penalty
dominates ($\lambda{=}0.5$, restricting beneficial moves) or
KPI weight is excessive ($\mu{=}0.5$, causing unstable early
training from noisy $\Delta\text{KPI}$ estimates).

The agent maximizes the expected discounted return
$G_t = \mathbb{E}\!\left[\sum_{k=0}^{\infty} \gamma^k
r_{t+k}\right]$ via a parameterized action-value network
$Q(s,a;\theta)$. A target network $Q_t(s,a;\theta^-)$ is
periodically synchronized to stabilize training. For each minibatch $\{(s_j, a_j, r_j, s_{j+1})\}_{j=1}^{N}$,
the temporal difference target is $y_j = r_j$ if $s_{j+1}$ is terminal and
$y_j = r_j + \gamma\, Q_t(s_{j+1}, a';\theta^-)$ otherwise,
with mean-squared loss
$\mathcal{L}(\theta) = \frac{1}{N}\sum_{j=1}^{N}
(y_j - Q(s_j, a_j;\theta))^2$.
Training uses Adam ($\eta{=}3{\times}10^{-4}$, $\gamma{=}0.99$,
batch size 64), a $10^5$-transition replay buffer with 2{,}000-step
warm-up, hard target updates every $\tau{=}1000$ steps, and
$\varepsilon$-greedy exploration decaying linearly from 1.0 to
0.05 over 50{,}000 steps. Gradient clipping (norm${=}10$) and
weight decay ($10^{-5}$) ensure stability across 2{,}000 episodes.

\subsection{Deep Q-Network}
The Q-network employs a hybrid CNN--BiLSTM--Attention
architecture designed to capture both spatial feature
correlations across warehouse state variables and long-range
temporal dependencies across decision steps. The state $s_t$
is first passed through a 1D-CNN, where each layer computes
$x_i^{(n)} = \mathrm{ReLU}(W_n \cdot x_{i-1} + b_n)$, to
extract local spatial patterns among co-located features such
as rack occupancy, stock levels, and picking frequency. The
resulting feature maps are fed into a BiLSTM,
producing forward and backward hidden representations
$\overrightarrow{H}_t$ and $\overleftarrow{H}_t$ that encode
both past operational trends and anticipated future demand
signals with standard gating mechanisms ($f_t$, $i_t$, $C_t$,
$o_t$).

To enable the agent to focus on the most operationally
critical time steps, an additive Bahdanau attention mechanism
\cite{abinaya2024enhanced} computes alignment scores
$e_i = v^T \tanh(W_a h_i + b_a)$, normalized via softmax
to obtain attention weights
$\alpha_i = \exp(e_i) / \sum_{k=1}^{T} \exp(e_k)$. The
context vector $c = \sum_{i=1}^{T} \alpha_i h_i$ is then
projected through a fully connected layer to produce Q-values
$Q(s_t, a;\theta)$ for all four zone actions. This selective
attention allows the network to weight warehouse state
transitions where demand shifts or congestion events
most strongly influence slotting decisions, directly supporting
the framework's objective of demand-aware zone assignment.

\subsection{Digital Twin Closed-Loop Integration}

As depicted in Fig.~\ref{fig1}, the agent-environment interaction follows a structured closed loop: the DQN agent issues a zone-assignment action $a_t$, which the action executor translates into FlexSim simulator commands via AAS. FlexSim advances the epoch and returns operational traces (picking durations, travel distances, queue states), from which the reward module computes $r_t$ and the next state $s_{t+1}$. The transition tuple $(s_t, a_t, r_t, s_{t+1})$ is stored in the response buffer for minibatch sampling and network updates. This perception decision execution loop supports online iterative policy refinement in a high-fidelity environment, while the AAS layer permits shadow evaluation of candidate slotting policies before any physical deployment.

\begin{table}[t]
\centering
\caption{FlexSim Digital Twin Warehouse Simulation Asset Configuration Parameters.}
\label{table1}
\resizebox{\columnwidth}{!}{%
\begin{tabular}{|l|c|l|c|}
\hline
\multicolumn{1}{|c|}{\textbf{Parameters}} & \textbf{Values} & \multicolumn{1}{c|}{\textbf{Parameters}} & \textbf{Values} \\ \hline
Inbound Queues Max Content & 100 & Racks & 16  \\ \hline
Outbound Queues Max Content & 100 & Zones & 4 \\ \hline
Inbound Conveyor Speed & 1.75 m\textbackslash{}s & Speed Picker Transporter & 1.25 m\textbackslash{}s \\ \hline
Outbound Conveyor Speed & 1.25 m\textbackslash{}s & Acceleration Picker Transporter & 1.75 m\textbackslash{}s \\ \hline
Operators & 2 & Speed Placer Transporter & 1.0 m\textbackslash{}s\\ \hline
 Sinks & 1  & Acceleration Placer Transporter & 1.25 m\textbackslash{}s \\ \hline
\end{tabular}%
}
\end{table}

\begin{figure} [t]
\centering
\includegraphics[width=8.5cm, height = 5cm]{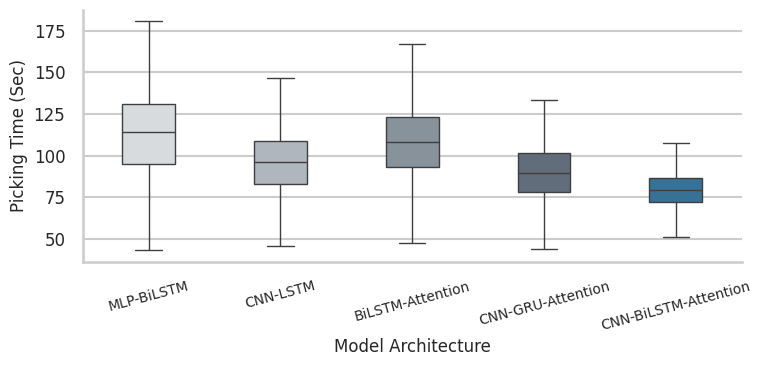}
\caption{Ablation study comparing average picking time across
Q-network architecture variants.}
\label{fig3}
\end{figure}

\begin{figure*}[h]
\centering
\includegraphics[width=18cm, height = 5.5cm]{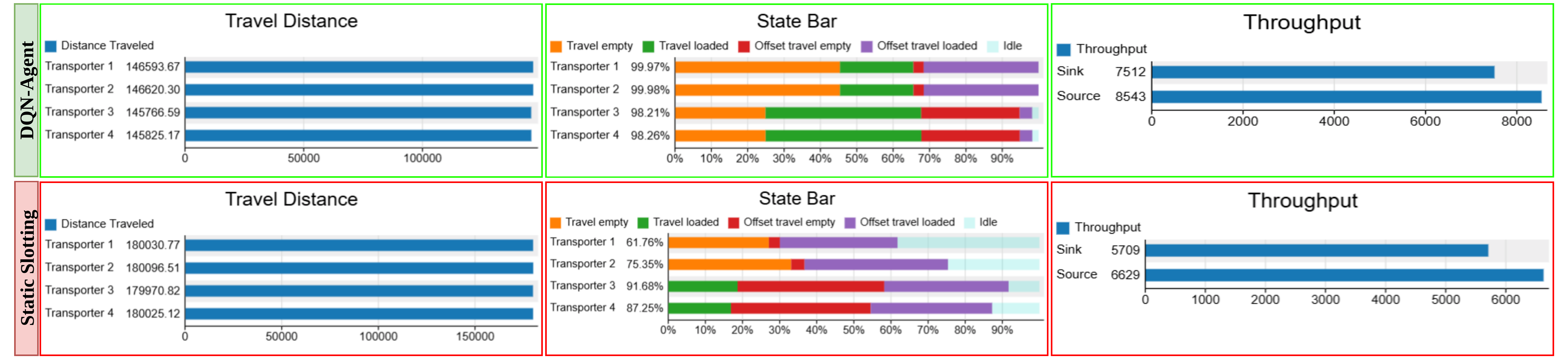}
\caption {FlexSim dashboard comparison between the proposed framework DQN-Agent and Static Slotting baseline across
per-transporter travel distance, utilization state distribution,
and warehouse throughput.}
\label{figure12}
\end{figure*}

\begin{table}[t]
\centering
\caption{Performance comparison averaged over 5 independent
runs with different random seeds.}
\label{table2}
\resizebox{\columnwidth}{!}{%
\begin{tabular}{|l|c|c|c|}
\hline
\textbf{Metrics} & \textbf{Static Slotting} &
\textbf{DQN-Agent} & \textbf{Changes (\%)} \\ \hline
Avg. Travel Distance (units) & 0.8203{\scriptsize$\pm$0.012}
& 0.6682{\scriptsize$\pm$0.018} &
\textbf{$\downarrow$18.54\%} \\ \hline
Avg. Picking Time (s) & 169.22{\scriptsize$\pm$2.41}
& 118.32{\scriptsize$\pm$3.15} &
\textbf{$\downarrow$29.66\%} \\ \hline
Throughput (orders/hour) & 23.35{\scriptsize$\pm$0.87}
& 31.62{\scriptsize$\pm$1.04} &
\textbf{$\uparrow$35.42\%} \\ \hline
\end{tabular}%
}
\end{table}

\subsection{Warehouse Logistics Digital Twin Environment}

The warehouse environment is instantiated as a high-fidelity 3D discrete-event simulation in FlexSim, serving as the interactive execution substrate for the DQN agent. The simulated facility encompasses four storage zones (A–D), each equipped with four double-sided racks, yielding 16 racks total, accommodating a 500-SKU product catalog. Zones are connected through navigable aisle networks that govern transporter routing and forklift trajectories, as visualized in the digital twin layout Fig. \ref{demo}.
Product flow is managed by four autonomous transporters: two Picker units (v = 1.25 m/s, a = 1.75 m/s²) responsible for retrieval to delivery movement, and two Placer units (v = 1.0 m/s, a = 1.25 m/s²) handling replenishment operations. Two inbound and two outbound queues (capacity: 100 units each) buffer product flows at the zone interfaces, while dedicated conveyors operating at 1.75 m/s inbound and 1.25 m/s outbound link storage areas to the dispatch sink. Two human operator agents and one sink node complete the material handling circuit, as summarized in Table \ref{table1}.
Each simulation epoch is triggered by the DQN agent's action executor, which translates zone assignment decisions into FlexSim object commands via the AAS interface layer. Upon epoch completion, the simulator emits operational traces picking durations, transporter trajectories, and queue states that are consumed by the reward computation module to close the agent-environment loop. This architecture enables safe shadow evaluation of slotting policies prior to any physical deployment, while maintaining continuous synchronization between the agent's learned model and the evolving simulated state.

\subsection{Experimental Setup}
All experiments are conducted on a workstation equipped with
an Intel Core i9-13900K CPU, 64\,GB RAM, and an NVIDIA
RTX 4090 GPU. The digital twin environment runs on FlexSim
25.1, with the DQN agent implemented in Python 3.10 using
PyTorch 2.1. The agent is trained for 2{,}000 episodes with each episode comprising a full order
cycle across the four-zone warehouse. Evaluation metrics are
averaged over several independent runs with distinct random
seeds to ensure reproducibility.

\section{Performance Analysis}
\label{sec4}

\subsection{Overall DQN Agent Performance}
As reported in Table~\ref{table2}, our framework demonstrates
consistent gains over the static slotting baseline (fixed zone
assignment by SKU category) across all three KPIs, averaged
over five runs with distinct random seeds. The agent reduces
average picking time by 29.66\% (169.22 to 118.32\,s),
improves throughput by 35.42\% (23.35 to 31.62 orders/hour),
and decreases travel distance by 18.54\%, reflecting its
learned ability to prioritize high-frequency items near
dispatch zones while internalizing spatial efficiency. Low
standard deviations across all metrics confirm policy
stability under varying initializations.

\subsection{Q-Network Architecture Ablation}
To validate our architectural choices, we compare the proposed
CNN-BiLSTM-Attention Q-network against four ablated variants,
as shown in Fig.~\ref{fig3}. The full architecture
achieves the lowest median picking time with the tightest
inter-quartile range, confirming that each component is essential: without the CNN, learning variance increases significantly, substituting GRU for BiLSTM reduces temporal modeling fidelity, and removing attention weakens the agent's ability to identify critical
state transitions, and replacing CNN with MLP yields the
poorest overall performance.

\subsection{Runtime Operational Analysis}
The FlexSim dashboard results in Fig.~\ref{figure12} provide
deeper operational insight into how our framework's slotting
decisions propagate through the warehouse system at runtime.
In terms of travel distance, the optimized configuration
reduces per-transporter travel by approximately 19\% across
all four units, a uniform improvement that confirms the agent
learns globally efficient zone assignments rather than
benefiting only specific transporters or routes. The transporter utilization results are particularly
revealing. Under static slotting, transporters spend up to
40\% of their time idle or repositioning without a load,
a direct consequence of poorly placed inventory forcing
long empty traversals. The DQN-Agent resolves this
inefficiency by aligning product locations with real-time
demand patterns, pushing all four transporters above 98\%
productive utilization. This near-elimination of idle time
represents a fundamental shift in how the warehouse
allocates its material handling resources, transforming
transporters from underutilized assets into continuous
productive units.

These efficiency gains compound naturally into throughput,
where the framework raises source throughput from 6{,}629
to 8{,}543 units and sink throughput from 5{,}709 to
7{,}512 units. This outcome underscores a key finding of
our work: intelligent slotting does not merely improve
individual metrics in isolation but creates a cascading
effect across the entire material flow pipeline, from
storage assignment through transporter routing to final
order delivery.

\section{Conclusion}\label{sec5}

This study introduces a dynamic slotting optimization framework based on DQN, integrated with a FlexSim Digital Twin for smart warehouse logistics. By unifying reinforcement learning with a high-fidelity simulation environment and the standardized AAS, the framework enables real-time, adaptive decision-making while maintaining consistent and interoperable asset modeling. The proposed approach is validated through extensive simulation, where the DQN agent consistently outperforms traditional static slotting methods by learning demand-aware and distance-efficient placement strategies. The results demonstrate significant improvements in operational efficiency, including faster picking processes, increased system throughput, and reduced travel distances. These gains are driven by the agent’s ability to prioritize high-demand items, minimize unnecessary transporter movements, and reduce idle and non-productive states.


\bibliographystyle{IEEEtran}

\bibliography{ICRA}

@article{alam2026ai,
  title={AI-enhanced Digital Twin systems for warehouse logistics optimization: A review of challenges with solutions and future directions},
  author={Alam, Md Mahinur and Ugli, Reimbaev Azizbek Nurkat and Tanha, Kanita Jerin and Jun, Taesoo},
  journal={ICT Express},
  year={2026},
  publisher={Elsevier}
}

@article{tanha2025flexsim,
  title={FlexSim-Based Digital Twin Framework for Smart Warehouse Monitoring and Performance Optimization},
  author={Tanha, Kanita Jerin and Alam, Md Mahinur and Reimbaev, Azizbek and Jun, Taesoo},
  journal={Proceedings of the Korean Institute of Communication Sciences Conference},
  pages={362--363},
  year={2025}
}

@article{abinaya2024enhanced,
  title={Enhanced image captioning using Bahdanau attention mechanism and heuristic beam search algorithm},
  author={Abinaya, Shanmugasundaram and Deepak, Mandava and Alphonse, A Sherly},
  journal={IEEE Access},
  volume={12},
  pages={100991--101003},
  year={2024},
  publisher={IEEE}
}

@article{jia2017dynamic,
  title={A dynamic logistic dispatching system with set-based particle swarm optimization},
  author={Jia, Ya-Hui and Chen, Wei-Neng and Gu, Tianlong and Zhang, Huaxiang and Yuan, Huaqiang and Lin, Ying and Yu, Wei-Jie and Zhang, Jun},
  journal={IEEE Transactions on Systems, Man, and Cybernetics: Systems},
  volume={48},
  number={9},
  pages={1607--1621},
  year={2017},
  publisher={IEEE}
}

@article{waubert2022dynamic,
  title={Dynamic storage location assignment in warehouses using deep reinforcement learning},
  author={Waubert de Puiseau, Constantin and Nanfack, Dimitri Tegomo and Tercan, Hasan and L{\"o}bbert-Plattfaut, Johannes and Meisen, Tobias},
  journal={Technologies},
  volume={10},
  number={6},
  pages={129},
  year={2022},
  publisher={MDPI}
}

@article{wu2024deep,
  title={Deep Reinforcement learning for task assignment and shelf reallocation in smart warehouses},
  author={Wu, Shao-Ci and Chiu, Wei-Yu and Wu, Chien-Feng},
  journal={IEEE Access},
  volume={12},
  pages={58915--58926},
  year={2024},
  publisher={IEEE}
}

@article{li2025reinforcement,
  title={A reinforcement learning framework for efficient task allocation among AGVs in smart warehouse},
  author={Li, Shaofeng and Zhao, Zejian and Wang, Di and Li, Ke and Liu, Gang and Wang, Quan},
  journal={IEEE Internet of Things Journal},
  volume={12},
  number={11},
  pages={16947--16961},
  year={2025},
  publisher={IEEE}
}

@article{islam2021novel,
  title={A novel particle swarm optimization-based grey model for the prediction of warehouse performance},
  author={Islam, Md Rakibul and Ali, Syed Mithun and Fathollahi-Fard, Amir Mohammad and Kabir, Golam},
  journal={Journal of Computational Design and Engineering},
  volume={8},
  number={2},
  pages={705--727},
  year={2021},
  publisher={Oxford University Press}
}

@article{hu2023dynamic,
  title={A dynamic integrated scheduling method based on hierarchical planning for heterogeneous AGV fleets in warehouses},
  author={Hu, Enze and He, Jianjun and Shen, Shuai},
  journal={Frontiers in neurorobotics},
  volume={16},
  pages={1053067},
  year={2023},
  publisher={Frontiers Media SA}
}

@article{cheng2023efficient,
  title={An efficient genetic algorithm for multi AGV scheduling problem about intelligent warehouse},
  author={Cheng, Wenlong and Meng, Wenjun},
  journal={Robotic Intelligence and Automation},
  volume={43},
  number={4},
  pages={382--393},
  year={2023},
  publisher={Emerald Publishing Limited}
}

@article{li2023double,
  title={Double-deck multi-agent pickup and delivery: Multi-robot rearrangement in large-scale warehouses},
  author={Li, Baiyu and Ma, Hang},
  journal={IEEE Robotics and Automation Letters},
  volume={8},
  number={6},
  pages={3701--3708},
  year={2023},
  publisher={IEEE}
}

\end{document}